Article

# Validation of the practicability of logical assessment formula for evaluations with inaccurate ground-truth labels: An application study on tumour segmentation for breast cancer

**Yongquan Yang[1,2,*], Hong Bu[1,3,*]**

[1] Institute of Clinical Pathology, West China Hospital, Sichuan University, Chengdu 610000, China
[2] Zhongju Flash Medical Technology Co., Ltd., Mianyang 621000, China
[3] Department of Pathology, West China Hospital, Sichuan University, Chengdu 610000, China
* **Corresponding authors:** Yongquan Yang, remy_yang@foxmail.com; Hong Bu, hongbu@scu.edu.cn

**Abstract:** The logical assessment formula (LAF) is a new theory proposed for evaluations with inaccurate ground-truth labels (IAGTLs) to assess the predictive models for artificial intelligence applications. However, the practicability of LAF for evaluations with IAGTLs has not yet been validated in real-world practice. In this paper, we applied LAF to two tasks of tumour segmentation for breast cancer (TSfBC) in medical histopathology whole slide image analysis (MHWSIA) for evaluations with IAGTLs. Experimental results and analysis show that the LAF-based evaluations with IAGTLs were unable to confidently act like usual evaluations with accurate ground-truth labels on the one easier task of TSfBC while being able to reasonably act like usual evaluations with AGTLs on the other more difficult task of TSfBC. These results and analysis reflect the potential of LAF applied to MHWSIA for evaluations with IAGTLs. This paper presents the first practical validation of LAF for evaluations with IAGTLs in a real-world application.

## 1. Introduction

The logical assessment formula (LAF) [1] has been proposed to achieve evaluations with inaccurate ground-truth labels (IAGTLs) to assess predictive models for various artificial intelligence applications. LAF aims to alleviate the situation of usual evaluations that need more or less accurate ground-truth labels (AGTLs) [2–6], and the situation of evaluations with IAGTLs that require the underlying true targets can be precisely defined [7–10]. LAF is suitable for evaluating the predicted targets of a predictive model in situations, where the underlying true targets are difficult to precisely define while multiple inaccurate targets that contain various information consistent with our prior knowledge about the underlying true target are available. Theoretical analysis of LAF revealed the practicability of LAF for evaluations with IAGTLs, which includes: 1) LAF can be applied for evaluations with IAGTLs on a more difficult task, able to act like usual strategies for evaluations with AGTLs reasonably; and 2) LAF can be applied for evaluations with IAGTLs simply from the logical point of view on an easier task, unable to act like usual strategies for evaluations with AGTLs confidently.

However, the revealed practicability of LAF for evaluations with IAGTLs has not yet been validated in real-world practice. In this paper, we aim to address this issue.

We applied LAF to tumour segmentation for breast cancer (TSfBC) in medical histopathology whole slide image analysis (MHWSIA). Based on two TSfBC tasks, we respectively evaluated two series of approaches with AGTLs-based usual strategy and IAGTLs-based LAF. Particularly, the two TSfBC tasks include a task that aims to segment tumours in HE-stained pre-treatment biopsy images and a task that aims to segment residual tumours in HE-stained post-treatment surgical resection images. According to pathology experts, the tumour segmentation task in HE-stained post-treatment surgical resection images is more difficult than the tumour segmentation task in HE-stained pre-treatment biopsy images. More details about the two tasks of TSfBC are available at Yang et al. [11]. A series of approaches chosen for evaluation include the baseline method (BaseLine) that directly learns from the inaccurate labels and various state-of-the-art methods for learning from inaccurate labels [12–19]. The other series of approaches chosen for evaluation include the approaches for the one series with one-step abductive multi-target learning (OSAMTL) [11] introduced. Extensive experiments were conducted, and corresponding results and analyses support that the practicability of LAF is valid in the case of TSfBC in MHWSIA, which reflect the potentials of LAF applied to MHWSIA for evaluations with IAGTLs.

The rest of the contents of this paper are structured as follows. In Section 2, we briefly review the related works. In Section 3, we give the detailed overview of LAF. In Section 3, we give the details of the implementation of LAF applied to TSfBC in MHWSIA. In Section 4, we conduct extensive experiments and analyse the corresponding results to validate the practicability of LAF in the case of TSfBC in MHWSIA. Finally, we conclude and discuss the whole paper in Section 5.

## 2. Related work

The aim of this paper is to validate the practicability of LAF [1], which is a new theory proposed for evaluations with IAGTLs, in real-world practice. Thus, evaluations with IAGTLs and LAF are related to this paper.

For evaluation with IAGTLs, two feasible types of methods have emerged. One is to firstly select some probably true targets from the inaccurate targets [9] within the given IAGTLs via probabilistic estimation, and then to achieve evaluations of unseen testing results by referring to the selected probably true targets [8,10]. The other is to achieve evaluations of unseen testing results by referring to the inaccurate targets [7] within the given IAGTLs with provided or estimated minimal rate of error corresponding to the true targets. Fundamentally, the assumption for these two types of methods is that there are true targets exist in the inaccurate targets represented by the given IAGTLs, which makes these two types of methods not suitable for the situation where the underlying true targets are difficult to be precisely defined or even do not exist, such as some applications in the field of MHWSIA [11,20,21].

To alleviate this issue, LAF [1] has been proposed. LAF has made two contributions to the literature of assessment for predictive models: 1) establishing a new theory for evaluations with IAGTLs, which does not need the assumption that there are true targets exist in the inaccurate targets represented by the given IAGTLs, and 2) offering a new addition to usual evaluations that require more or less AGTLs [2–6] as well as some existing methods for evaluations with IAGTLs [7–10]. More detailed overview of LAF is provided in Section 3.

## 3. Overview of logical assessment formula

As the purpose of this paper is to validate the practicability of LAF for evaluations with IAGTLs in real-world practice, LAF is highly related to this work. In this section, we briefly present an overview of LAF. More details about LAF and its principles for evaluations with IAGTLs are provided at Yang [1].

### 3.1. Formation and usage of LAF

The formation of LAF [1] can be formally denoted as

$$LAF \begin{cases} inputs: \begin{cases} t \\ \tilde{t} = \{\tilde{t}_1, \cdots, \tilde{t}_m\} \end{cases} \\ PC \begin{cases} LF = LogicalFactNarrate(\tilde{t}; p^{LFN}) \\ LC = LogicalConsistencyEstimate(t, LF; p^{LCE}) \\ LAM = LogicalAssessmentMetricBuild(LC; p^{LAM}) \end{cases} \\ output: LAM = \{LAM_1, \cdots, LAM_w\} \end{cases} \quad (1)$$

Specifically, given the predicted target ($t$) for the underlying true targets, which are difficult to precisely define, and multiple inaccurate targets ($\tilde{t} = \{\tilde{t}_1, \cdots, \tilde{t}_m\}$) that contain various information consistent with our prior knowledge about the underlying true target, we can obtain, via the processing components of LAF ($LAF: PC$), a series of logical assessment metrics ($LAM$) for evaluations of the given predicted target ($t$) compared with the underlying true target. $LAF: PC$ is constituted by three components, including logical fact narration, logical consistency estimation, and logical assessment metric build.

Narrating logical facts ($LF$) from the input multiple inaccurate targets ($\tilde{t}$), the logical fact narration component produces a list of qualitative descriptions ($LF = \{LF_1, \ldots, LF_f\}$) that logically represent the facts contained in the given multiple inaccurate targets ($\tilde{t}$). Estimating the logical consistencies ($LC$) between the input predicted target ($t$) and the narrated logical facts ($LF$), the logical consistency estimation component generates a list of qualitative descriptions ($LC = \{LC_1, \ldots, LC_u\}$) that logically represent the consistencies between the given predicted target ($t$) and the narrated $LF$. Producing a series of logical assessment metrics ($LAM$) based on the estimated logical consistencies ($LC$) between the input predicted target ($t$) and the narrated logical facts ($LF$), the logical assessment metric build component outputs a series of abstractly formalised metrics ($LAM = \{LAM_1, \cdots, LAM_w\}$) that are derived from the qualitative descriptions of the estimated $LC$ to represent the evaluations of the predicted target ($t$) compared with the underlying true target.

Formally, the usage of LAF can be denoted as

$$LAM = LAF: PC(t, \tilde{t}; \{p^{LFN}, p^{LCE}, p^{LAM}\}) = \{LAM_1, \cdots, LAM_w\} \quad (2)$$

Each $p^*$ of Equation (2) denotes the hyperparameters corresponding to the implementation of respective expression of $LAF: PC$.

In summary, the outline of LAF for evaluations with IAGTLs is shown as **Figure 1**.

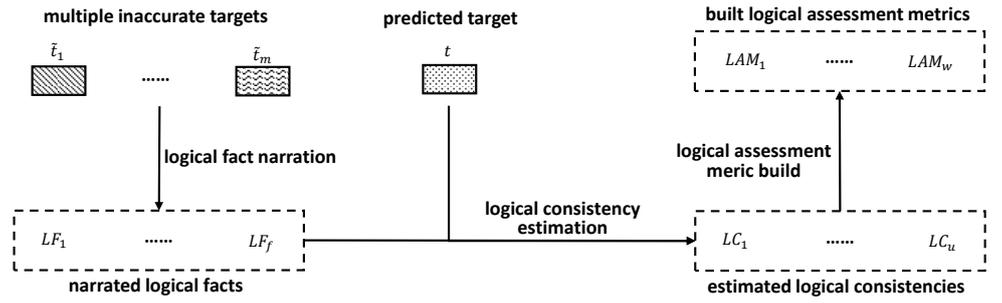

**Figure 1.** Outline of LAF for evaluations with IAGTLs [1,21,22].

## 3.2. LAF-based method performance evaluation

The LAF-based method performance evaluation (LMP) strategy is to estimate the effectiveness of a method for addressing a task. As the method and the task should be specifically given in advance, LMP is task-specific (ts) and method-specific (ms). The input of LMP is a series of task-specific and method-specific logical assessment metrics ($LAM_{ts,ms}$). The output of LMP is some method performances ($LMP_{ts,ms}$), which are respectively quantized in the range [0,1], to reflect the superiorities of the given specific method for addressing a specific task. As a result, the processing procedure of LMP can be formally expressed as

$$LMP_{ts,ms} = LogicalMethodPerfEval\left(LAM_{ts,ms}; p^{LMP_{ts,ms}}\right)$$
$$= \{LMP_{ts,ms,1}, \cdots, LMP_{ts,ms,v}\}, Val(LMP_{ts,ms,v}) \in [0,1] \quad (3)$$

here, $p^{LMP_{ts,ms}}$ denotes the hyperparameters for implementation of Equation (3) and $Val(*)$ denotes the value of $*$.

## 3.3. Practicability of LAF

The practicability of LAF is as follows:

- Practicability 1. LAF can be applied for evaluations with IAGTLs on a more difficult task, able to act like usual strategies for evaluations with AGTLs reasonably.
- Practicability 2. LAF can be applied for evaluations with IAGTLs simply from the logical point of view on an easier task, unable to act like usual strategies for evaluations with AGTLs confidently.

## 4. LAF Applied to tumour segmentation for breast cancer

In this section, we apply LAF to two tasks of tumour segmentation for breast cancer (TSfBC) in medical histopathology whole slide image analysis (MHWSIA) for evaluations with inaccurate ground-truth labels (IAGTLs). Since it is indeed difficult to accurately annotate the true targets for both of the two tasks [11], LAF-based evaluations with IAGTLs just provide a good alternative for this situation. In Section 4.1, we briefly describe the two tasks of TSfBC. In Section 4.2, we give descriptions of the settings for the application of LAF to TSfBC. In Section 4.3, we provide the details of the implementations of LAF applied to TSfBC.

### 4.1. Tumour segmentation for breast cancer

The two tasks of TSfBC include a task that aims to segment tumours in HE-

stained pre-treatment biopsy images and a task that aims to segment residual tumours in HE-stained post-treatment surgical resection images. Referring to additional suggestions from pathology experts, we here claim that the tumour segmentation task in HE-stained post-treatment surgical resection images is more difficult than the tumour segmentation task in HE-stained pre-treatment biopsy images. More details about challenges and difficulty comparisons for the two tasks of TSfBC are available at Yang et al. [11].

### 4.2. Application settings

Since our main purpose in this application is to apply LAF to the two tasks of TSfBC for evaluations with IAGTLs, we focus more on the settings required by LAF instead of the details of the specific methods for addressing the two tasks.

#### 4.2.1. Inputs of LAF

The outline of the inputs of LAF applied to TSfBC is shown as **Figure 2**. Due to the fact that the underlying true targets for the two tasks of TSfBC are difficult to precisely define, we set up the two tasks as problems of learning from inaccurate (noisy) labels [23,24]. Testing samples with IAGTLs provided by pathology experts for the two tasks of TSfBC are shown in the middle of **Figure 2**. In the middle of **Figure 2**, IAGTLs (1) include many non-tumour areas as tumour areas while IAGTLs (2) exclude many tumour areas as non-tumour areas, which indicates that preparing IAGTLs requires much less labour. Two types of inaccurate targets corresponding to the testing samples are extracted from the given IAGTLs via one-step abductive logical reasoning [11]. Examples of the two types of inaccurate targets extracted corresponding to the testing samples are shown on the left of **Figure 2**. The predicted targets corresponding to the testing samples are obtained via an image semantic segmentation model trained with methods for learning from inaccurate labels, which will be discussed later in Section 4.2.2–3. Examples of the predicted targets corresponding to the testing samples are shown on the right of **Figure 2**.

Here, we omitted the details of extracting the two types of inaccurate targets since our main purpose in this section is to implement the application of LAF to TSfBC for evaluations with IAGTLs. But we claim that the extracted two types of inaccurate targets contain information consistent with our prior knowledge about the underlying true targets, referring to the one-step abductive logical reasoning presented in our previous work [11]. More specifically, the extracted targets (1) ($\tilde{t}_{TSfBC,1}$) can maintain high recall of the underlying true targets of TSfBC, and the extracted targets (2) ($\tilde{t}_{TSfBC,2}$) can maintain high precision of the underlying true targets of TSfBC. In summary, the two types of inaccurate targets can be extracted based on logical reasoning, and more details can be found in our previous work [11]. As a result, we denote the multiple inaccurate targets that contain various information consistent with our prior knowledge about the underlying true targets of TSfBC by

$$\tilde{t}_{TSfBC} = \left\{ \tilde{t}_{TSfBC,1}, \tilde{t}_{TSfBC,2} \right\} \tag{4}$$

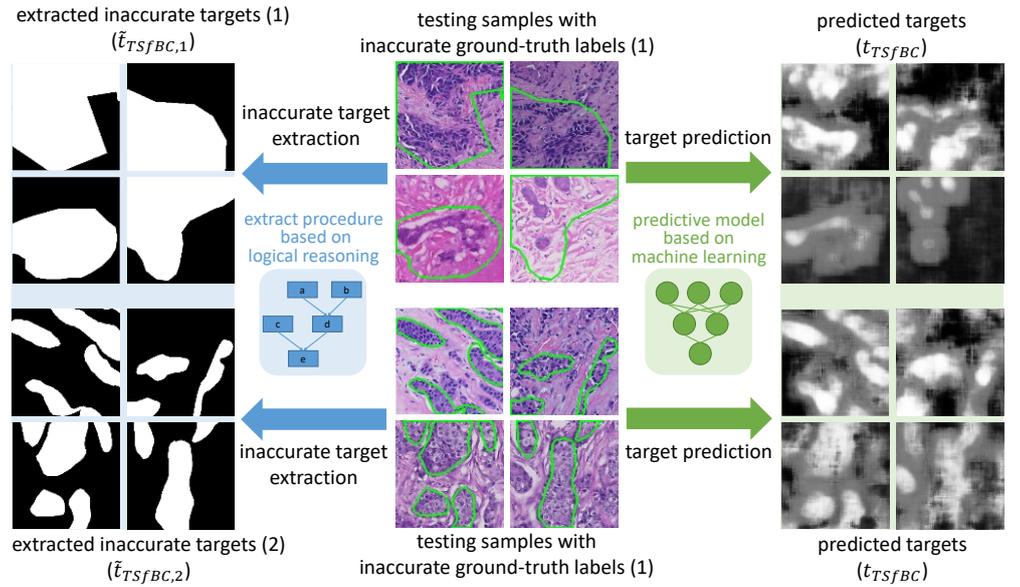

**Figure 2.** Outline of the settings for the inputs of LAF applied to TSfBC. Middle: testing samples with inaccurate ground-truth labels (IAGTLs); Left: inaccurate targets corresponding to testing samples; Right: predicted targets corresponding to testing samples.

### 4.2.2. Image semantic segmentation model

The base image semantic segmentation model (ISSM) for the predicted targets corresponding to the testing samples for the two tasks of TSfBC is a symmetric deep convolutional neural network (DCNN) that was built for the task of H. pylori segmentation [20,21]. The symmetric image semantic segmentation architecture was implemented by referring to the most commonly used fully convolutional network [25], which is representative of fully convolutional network-based solutions and has inspired various other solutions achieving state-of-the-art performances in image semantic segmentation. Another reason for choosing this architecture for implementing the base ISSM is processing efficiency, as the two tasks of TSfBC are defined on whole slide images, the dimensions of which are very large. More details about the architecture of the symmetric DCNN can be found in Yang et al. [21]. We let $\{cnn_l\}_{l=0}^{X}$ denote the transformation for each of the $X$ layers in the built base DCNN, $\{w_l\}_{l=0}^{X}$ denote the parameters of $\{cnn_l\}_{l=0}^{X}$, and $p^{DCNN}$ denote the hyperparameters for the optimisation of $\{w_l\}_{l=0}^{X}$. We assume that the input of the built-in DCNN (an image instance) is $I$ and the output of the built base DCNN (a predicted target corresponding to the input image instance $I$) is $t_{TSfBC}$. With all these denotations and assumptions, we can express the image semantic segmentation model (ISSM) for the two tasks of TSfBC by

$$t_{TSfBC} = ISSM(I; \{DCNN, p^{DCNN}\}) \tag{5}$$

$$DCNN = \{\{cnn_l\}_{l=0}^{X}, \{w_l\}_{l=0}^{X}\} \tag{6}$$

Note, in practice, $p^{DCNN}$ can be a designated method of learning from inaccurate labels based on deep learning, since we set up the two tasks of TSfBC as problems of learning from noisy labels.

### 4.2.3. Methods of learning from inaccurate labels

In addition to the baseline method (BaseLine) that directly learns from the inaccurate labels, various state-of-the-art methods for learning from inaccurate labels, including Forward, Backward [12], Boost-Hard, Boost-Soft [13,14], D2L [15], SCE [16], Peer [17], DT-Forward [18], and NCE-SCE [19], are also chosen to designate the hyperparameter $p^{DCNN}$ for experimental investigations. These state-of-the-art methods are chosen due to their flexibility to be applied to the situation, where no clean dataset is available, the targeted object cannot be precisely defined, and any of the given inaccurate labels cannot be confidently regarded as probably true targets. In addition, these state-of-the-art methods, combined with an improved version of one-step abductive multi-target learning (OSAMTL) [11], were also chosen to designate the hyperparameter $p^{DCNN}$ for experimental investigations. We set the hyperparameters of these approaches as suggested by the corresponding papers. We denote the designated $p^{DCNN}$ by the method-specific (ms) $p_{ms}^{DCNN}$. As a result, we rewrite the formulation of the image semantic segmentation model for the two tasks of TSfBC by

$$t_{TSfBC,ms} = ISSM(I; \{DCNN, p_{ms}^{DCNN}\}),$$
$$ms \in \{BaseLine, \cdots, NCE - SCE, BaseLine\_OSAMTL, \cdots, NCE - SCE\_OSAMTL\} \quad (7)$$

## 4.3. Implementation of LAF applied to TSfBC

On the basis of LAF overviewed in Section 3 and the application settings required by LAF to be carried out, we provide an implementation of LAF suitable to be applied for evaluations with IAGTLs on TSfBC.

### 4.3.1. Implementation of task-specific LAF

We implement a task-specific LAF that is suitable for evaluations with IAGTL on TSfBC. Referring to **Figure 1**, the outline for the application of LAF to TSfBC is summarized as **Figure 3**.

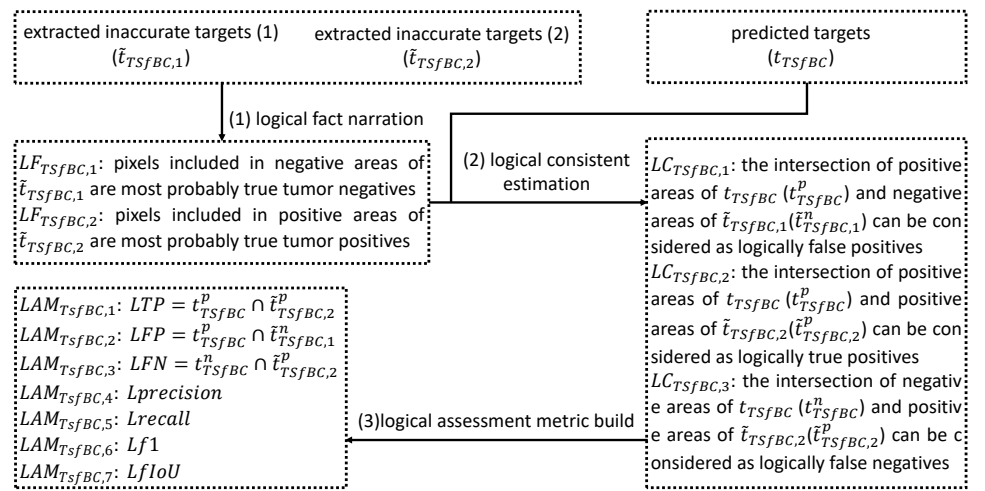

**Figure 3.** Outline for the application of LAF to TSfBC.

Referring to Equation (1) and letting $ts = TSfBC$ and $m = 2$, we can denote the task-specific LAF that is suitable for evaluations with IAGTL on TSfBC as

$$LAF \begin{cases} inputs: \begin{cases} t_{TSfBC} \\ \tilde{t}_{TSfBC} = \{\tilde{t}_{TSfBC,1}, \tilde{t}_{TSfBC,2}\} \end{cases} \\ PC \begin{cases} LF_{TSfBC} = LogicalFactNarrate(\tilde{t}_{TSfBC}; p_{TSfBC}^{LFN}) \\ LC_{TSfBC} = LogicalConsistencyEstimate(t_{TSfBC}, LF_{TSfBC}; p_{TSfBC}^{LCE}) \\ LAM_{TSfBC} = LogicalAssessmentMetricBuild(LC_{TSfBC}; p_{TSfBC}^{LAM}) \end{cases} \\ outputs: LAM_{TSfBC} \end{cases} \quad (8)$$

We need to clearly define each $p_{TSfBC}^*$ of respective processing component for the implementation of task-specific LAF, regarding to the inherent characteristics of TSfBC.

(1) Logical facts narration

On the basis of the claim that the inaccurate targets $\tilde{t}_{TSfBC} = \{\tilde{t}_{TSfBC,1}, \tilde{t}_{TSfBC,2}\}$ in Section 4.2.1 contain information consistent with our prior knowledge about the underlying true target, and the given inaccurate target $\tilde{t}_{TSfBC,1}$ can keep high recall of the underlying true target of TSfBC and the given inaccurate target $\tilde{t}_{TSfBC,2}$ can keep high precision of the underlying true target of TSfBC, we introduce two reasonings (Reasoning 1 and Reasoning 2). The validity of the two derived reasonings are respectively proved by Proof-R1 and Proof-R2 which are provided in Supplementary.

**Reasoning 1.** *If $\tilde{t}_{TSfBC,1}$ is given, then pixels included in negative areas of $\tilde{t}_{TSfBC,1}$ are most probably true tumour negatives.*

**Reasoning 2.** *If $\tilde{t}_{TSfBC,2}$ is given, then pixels included in positive areas of $\tilde{t}_{TSfBC,2}$ are most probably true tumour positives.*

Referring to Equation (8) and using Reasonings 1 and 2 as $p_{TSfBC}^{LFN}$, we implement the $LogicalFactNarrate$, which narrates two logical facts from $\tilde{t}_{TSfBC}$, as follows

$$LF_{TSfBC} = LogicalFactNarrate(\tilde{t}_{TSfBC}; \{Reasoning\ 1, Reasoning\ 2\})$$
$$= \begin{cases} LogicalFactNarrate(\tilde{t}_{TSfBC,1}; \{Reasoning\ 1\}), \\ LogicalFactNarrate(\tilde{t}_{TSfBC,2}; \{Reasoning\ 2\}) \end{cases} \quad (9)$$
$$= \{LF_{TSfBC,1}, LF_{TSfBC,2}\}$$

Details of the narrated two logical facts are provided in **Table 1**.

**Table 1.** Details of the narrated logical facts.

| Narrated Logical Facts |
| --- |
| $LF_{TSfBC,1}$: pixels included in negative areas of $\tilde{t}_{TSfBC,1}$ are most probably true tumour negatives |
| $LF_{TSfBC,2}$: pixels included in positive areas of $\tilde{t}_{TSfBC,2}$ are most probably true toumour positives |

(2) Logical consistency estimation

On the basis of the prediction of the image semantic segmentation model for tumour segmentation for breast cancer ($t_{TSfBC}$) in Section 4.2.2 and the two narrated logical facts $LF_{TSfBC} = \{LF_{TSfBC,1}, LF_{TSfBC,2}\}$, we introduce two reasonings (Reasoning 3 and Reasoning 4). The validity of the two derived reasonings are respectively proved by Proof-R3 and Proof-R4 which are provided in Supplementary.

**Reasoning 3.** *If $t_{TSfBC}$ is given and $LF_{TSfBC,1}$ is given, then the intersection of pixels of $t_{TSfBC}$ that are predicted as tumour positives ($t_{TSfBC}^p$) and pixels included in negative areas of $\tilde{t}_{TSfBC,1}$ ($\tilde{t}_{TSfBC,1}^n$) can be considered as logically false positives.*

**Reasoning 4.** *If $t_{TSfBC}$ is given and $LF_{TSfBC,2}$ is given, then the intersection of pixels of $t_{TSfBC}$ that are predicted as tumour positives ($t_{TSfBC}^p$) and pixels included in positive areas of $\tilde{t}_{TSfBC,2}$ ($\tilde{t}_{TSfBC,2}^p$) can be considered as logically true positives, and the intersection of pixels of $t_{TSfBC}$ that are predicted as tumour negatives ($t_{TSfBC}^n$) and pixels included in positive areas of $\tilde{t}_{TSfBC,2}$ ($\tilde{t}_{TSfBC,2}^p$) can be considered as logically false negatives.*

Referring to Equation (8) and using Reasonings 3 and 4 as $p_{TSfBC}^{LCE}$, we implement the *LogicalConsistencyEstimate*, which estimates three logical consistencies between $t_{TSfBC}$ and $LF_{TSfBC}$, as follows

$$
\begin{aligned}
LC_{TSfBC} &= LogicalConsistencyEstimate\left(t_{TSfBC}, LF_{TSfBC}; \begin{Bmatrix} Reasoning\ 3, \\ Reasoning\ 4 \end{Bmatrix}\right) \\
&= \begin{Bmatrix} LogicalConsistencyEstimate(t_{TSfBC}, LF_{TSfBC,1}; \{Reasoning\ 3\}), \\ LogicalConsistencyEstimate(t_{TSfBC}, LF_{TSfBC,2}; \{Reasoning\ 4\}) \end{Bmatrix} \\
&= \{LC_{TSfBC,1}, LC_{TSfBC,2}, LC_{TSfBC,3}\}
\end{aligned} \tag{10}
$$

Details of the estimated three logical consistencies are provided in **Table 2**.

**Table 2.** Details of the estimated logical consistencies.

| Estimated Logical Consistencies |
| --- |
| $LC_{TSfBC,1}$: the intersection of $t_{TSfBC}^p$ and $\tilde{t}_{TSfBC,1}^n$ can be considered as logically false positives |
| $LC_{TSfBC,2}$: the intersection of $t_{TSfBC}^p$ and $\tilde{t}_{TSfBC,2}^p$ can be considered as logically true positives |
| $LC_{TSfBC,3}$: the intersection of $t_{TSfBC}^n$ and $\tilde{t}_{TSfBC,2}^p$ can be considered as logically false negatives |

(3) Logical assessment metric build

Based on the estimated $LC_{TSfBC}$, referring to Equation (8) and using usual definitions for assessment of image semantic segmentation as $p_{TSfBC}^{LAM}$, we implement *LogicalAssessmentMetricBuild* to abstractly formalize a series of logical assessment metrics, which can be expressed as

$$
\begin{aligned}
LAM_{TSfBC} &= LogicalAssessmentMetricBuild\left(LC_{TSfBC}; \begin{Bmatrix} TP, FP, FN, \\ precision, recall, \\ f1, fIoU \end{Bmatrix}\right) \\
&= \begin{Bmatrix} LAM_{TSfBC,1}, LAM_{TSfBC,2}, LAM_{TSfBC,3}, \\ LAM_{TSfBC,4}, LAM_{TSfBC,5}, LAM_{TSfBC,6}, LAM_{TSfBC,7} \end{Bmatrix}.
\end{aligned} \tag{11}
$$

Details of the built logical assessment metrics are provided in **Table 3**.

**Table 3.** Details of the build logical assessment metrics.

| Built Logical Assessment Metrics |
| --- |
| $LAM_{TSfBC,1}$: $LTP = t_{TSfBC}^p \cap \tilde{t}_{TSfBC,2}^p$ |
| $LAM_{TSfBC,2}$: $LFP = t_{TSfBC}^p \cap \tilde{t}_{TSfBC,1}^n$ |
| $LAM_{TSfBC,3}$: $LFN = t_{TSfBC}^n \cap \tilde{t}_{TSfBC,2}^p$ |
| $LAM_{TSfBC,4}$: $Lprecision = \frac{LTP}{LTP+LFP}$ |

**Table 3.** (*Continued*).

| Built Logical Assessment Metrics |
| --- |
| $LAM_{TsfBC,5}$: $Lrecall = \frac{LTP}{LTP+LFN}$ |
| $LAM_{TsfBC,6}$: $Lf1 = \frac{2 \times Lprecision \times Lrecall}{Lprecision+Lrecall}$ |
| $LAM_{TsfBC,7}$: $LfIoU = \frac{LTP}{LTP+LFP+LFN}$ |

(4)  Result

Based on the implemented task specific LAF ($LAF_{TsfBC}$), we can get a series of abstractly formalized metrics that that are suitable for evaluations with IAGTL on TSfBC. As a result, referring to Equations (8) and (2), the abstractly formalized metrics can be denoted by

$$LAM_{TsfBC} = LAF : PC\left(t_{TSfBC}, \tilde{t}_{TSfBC}; \left\{p_{TSfBC}^{LFN}, p_{TSfBC}^{LCE}, p_{TSfBC}^{LAM}\right\}\right)$$
$$= \left\{LAM_{TsfBC,1}, \cdots, LAM_{TsfBC,7}\right\} \tag{12}$$

### 4.3.2. Implementation of method-specific LAF

Regarding the various methods of learning from noisy labels referred to Section 4.2.3, we can designate $t_{TSfBC}$ to be associated with a specific method of learning from noisy labels. With the $t_{TSfBC}$ designated to be associated with a specific method of learning from noisy labels, we can transform the abstractly formalised $LAM_{TSfBC}$ into quantitative values of assessment to implement the method-specific LAF for evaluations with IAGTL on TSfBC. Referring to Equation (12) and letting $ms$ be a specific method of learning from noisy labels, the transformed quantitative values of assessment can be denoted by

$$LAM_{TSfBC,ms} = LAF : PC\left(t_{TSfBC,ms}, \tilde{t}_{TSfBC}\right)$$
$$= \left\{LAM_{TSfBC,ms,1}, \cdots, LAM_{TSfBC,ms,7}\right\}, ms \in \{BaseLine, Forward, \cdots, OSAMTL\}. \tag{13}$$

### 4.3.3. Implementation of LAF based method performance evaluation

Based on the transformed quantitative values of assessment for evaluations with IAGTL on TSfBC ($LAM_{TSfBC,ms}$), and referring to Equations (13) and (3), we can derive LAF based method performance (LMP). For a simple implementation of LMP, we set the hyper-parameters $p^{LMP_{TSfBC,ms}}$ for implementation of $LogicalMethodPerfEval$ by 'selecting the metric of overall performance (SMOP)', which can be expressed as

$$LMP_{TSfBC,ms} = LogicalMethodPerfEval\left(LAM_{TSfBC,ms}, 'SMOP'\right)$$
$$= \left\{LAM_{TSfBC,ms,6}, LAM_{TSfBC,ms,7}\right\} \tag{14}$$

## 5. Verification for practicability of LAF

On the basis of the application of LAF to two tasks of tumour segmentation for breast cancer (TSfBC) in medical histopathology whole slide image analysis (MHWSIA) presented in Section 4, in this section, we conduct experiments and give corresponding analysis to further verify the practicability of LAF for evaluations with inaccurate ground-truth labels (IAGTLs).

## 5.1. Preliminary

### 5.1.1. Overall design

Referring to the summarised practicability of LAF, we consider two key points that need to be experimentally verified to better realise the pros and cons of LAF. The two key points include: 1) on a more difficult task, LAF is able to act like usual strategies for evaluations with AGTLs reasonably; and 2) on an easier task, LAF is unable to act like usual strategies for evaluations with AGTLs confidently.

To verify these two key points, we first conduct experiments that employ LAF to produce evaluations of various methods for learning from inaccurate labels with IAGTLs and experiments that employ the usual strategy (US) to produce evaluations of various methods for learning from inaccurate labels with AGTLs, on the two tasks of tumour segmentation for breast cancer (**Figure 2**). For each of the two tasks, we conduct two series of experiments, including a number of state-of-the-art methods [12–19] for learning from inaccurate labels and their respective combinations with an improved version of OSAMTL [11]. As the previous work [11] has confirmed the advantages of the improved OSAMTL series compared with the state-of-the-art series [12–19] using US-based evaluations with AGTLs, we can compare the results of the improved OSAMTL series with the results of the state-of-the-art series using LAF-based evaluations with IAGTLs to observe whether the LAF-based evaluations with IAGTLs can maintain the advantages of the improved OSAMTL series.

According to the two key points that need to be verified, specifically, we have two expectations in advance: 1) Evaluations of LAF with IAGTLs can show the advantages of the improved OSAMTL series compared with the state-of-the-art series, just being able to reasonably act like evaluations of US with AGTLs on the task of tumour segmentation in HE-stained post-treatment surgical resection images, which is more difficult; 2) Evaluations of LAF with IAGTLs cannot show the advantages of the improved OSAMTL series compared with the state-of-the-art series, just being unable to confidently act like evaluations of US with AGTLs on the task of tumour segmentation in HE-stained pre-treatment biopsy images, which is easier.

### 5.1.2. Data preparation

For evaluations with IAGTLs using LAF on the task of tumour segmentation in HE-stained pre-treatment biopsy images, we prepared 248 image patches with IAGTLs (1) corresponding to $\tilde{t}_{TSfBC,1}$ and 36 image patches with IAGTLs (2) corresponding to $\tilde{t}_{TSfBC,2}$. For evaluations with AGTLs using US on the task of tumour segmentation in HE-stained pre-treatment biopsy images, we prepared 158 image patches with corresponding AGTLs.

For evaluations with IAGTLs using LAF on the task of tumour segmentation in HE-stained post-treatment surgical resection images, we prepared 736 image patches with IAGTLs (1) corresponding to $\tilde{t}_{TSfBC,1}$ and 358 image patches with IAGTLs (2) corresponding to $\tilde{t}_{TSfBC,2}$. For evaluations with AGTLs using US on the task of tumour segmentation in HE-stained pre-treatment biopsy images, we prepared 242 image patches with corresponding AGTLs.

The image patches prepared for experiments were cropped at $10 \times$ magnification of some digital whole slide images, and the size of each cropped image patch was 256

× 256 pixels (width × height). Some examples of the image patches prepared for evaluations with IAGTLs or AGTLs on the two tasks are provided in **Figure 4**. From **Figure 4**, we can note that the preparation of the image patches for evaluations with IAGTLs is much less labour intensive than the preparation of the image patches for evaluations with AGTLs.

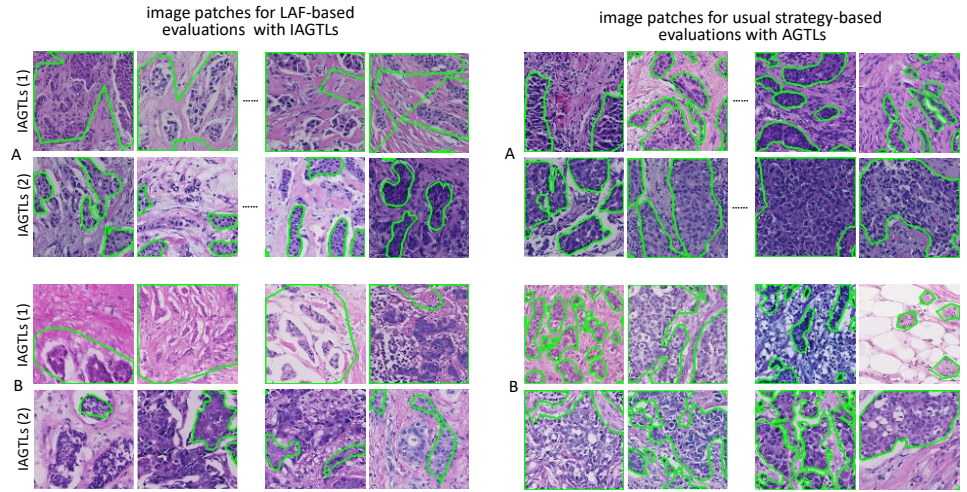

**Figure 4.** Examples of the image patches prepared for evaluations with IAGTLs or AGTLs on the two tasks of TSfBC. **(A)** the task of tumour segmentation in HE-stained pre-treatment biopsy images; **(B)** the task of tumour segmentation in HE-stained post-treatment surgical resection images.

### 5.1.3. Experimental settings

All of our experiments were performed on an Intel Core Xeon E5-2630 v4s with a memory capacity of 128GB and eight NVIDIA GTX 1080Ti GPUs. Our developing environment is based on Tensorflow 1.10.1 and Python 3.5. More detailed experimental settings for training the image semantic segmentation model with the two series of methods of learning from inaccurate labels to produce the predictions can be found in our previous work [11].

## 5.2. Results of LAF-based evaluations with IAGTLs

Referring to the implementations of LAF applied on TSfBC presented in Section 4, the LAM and LMP results of LAF-based evaluations with IAGTLs for various methods of learning from inaccurate labels for the tumour segmentation in HE-stained pre-treatment biopsy images and the tumour segmentation in HE-stained post-treatment surgical resection images are respectively shown in **Tables 4** and **5**.

**Table 4.** LAF-based evaluations with IAGTLs on the task of tumour segmentation in HE-stained pre-treatment biopsy images.

| Solution | LAM | | | | | LMP | |
|---|---|---|---|---|---|---|---|
| | LTP | LFP | LFN | Lprecision | Lrecall | Lfl | LfloU |
| BaseLine | 17,619 | 6956 | 1698 | 71.69 | 91.21 | 80.28 | 67.06 |
| Forward | 17,455 | 5680 | 1861 | 75.45 | 90.37 | 82.24 | 69.83 |
| Backward | 15,175 | 7032 | 4141 | 68.33 | 78.56 | 73.09 | 57.59 |
| Boost-Hard | 17,497 | 7104 | 1820 | 71.12 | 90.58 | 79.68 | 66.22 |
| Boost-Soft | 15,685 | 6564 | 3631 | 70.50 | 81.20 | 75.47 | 60.61 |
| D2l | 17,506 | 7697 | 1811 | 69.46 | 90.62 | 78.64 | 64.80 |
| SCE | 16,627 | 5601 | 2690 | 74.80 | 86.07 | 80.04 | 66.73 |
| Peer | 17,669 | 6775 | 1648 | 72.28 | 91.47 | 80.75 | 67.72 |
| DT-Forward | 16,731 | 5814 | 2586 | 74.21 | 86.61 | 79.93 | 66.58 |
| NCE-SCE | 16,901 | 6605 | 2415 | 71.90 | 87.50 | 78.94 | 65.20 |
| BaseLine_OSAMTL | 15,428 | 4165 | 3888 | 78.74 | 79.87 | 79.30 | 65.70 |
| Forward_OSAMTL | 14,132 | 3282 | 5184 | 81.15 | 73.16 | 76.95 | 62.54 |
| Backward_OSAMTL | 15,414 | 3816 | 3902 | 80.16 | 79.8 | 79.98 | 66.63 |
| Boost-Hard_OSAMTL | 14,928 | 3812 | 4389 | 79.66 | 77.28 | 78.45 | 64.54 |
| Boost-Soft_OSAMTL | 15,511 | 5198 | 3805 | 74.9 | 80.3 | 77.51 | 63.27 |
| D2l_OSAMTL | 15,220 | 4267 | 4097 | 78.1 | 78.79 | 78.45 | 64.54 |
| SCE_OSAMTL | 14,982 | 4264 | 4334 | 77.84 | 77.56 | 77.7 | 63.54 |
| Peer_OSAMTL | 14,637 | 4182 | 4680 | 77.78 | 75.77 | 76.76 | 62.29 |
| DT-Forward_OSAMTL | 14,675 | 2956 | 4641 | 83.23 | 75.97 | 79.44 | 65.89 |
| NCE-SCE_OSAMTL | 14,238 | 3993 | 5078 | 78.1 | 73.71 | 75.84 | 61.08 |

**Table 5.** LAF-based evaluations with IAGTLs on the task of tumour segmentation in HE-stained post-treatment surgical resection images.

| Solution | LAM | | | | | LMP | |
|---|---|---|---|---|---|---|---|
| | LTP | LFP | LFN | Lprecision | Lrecall | Lfl | LfloU |
| BaseLine | 16,131 | 7863 | 4525 | 67.23 | 78.09 | 72.26 | 56.56 |
| Forward | 14,933 | 7440 | 5723 | 66.75 | 72.29 | 69.41 | 53.15 |
| Backward | 15,196 | 8983 | 5460 | 62.85 | 73.57 | 67.79 | 51.27 |
| Boost-Hard | 15,829 | 8878 | 4826 | 64.07 | 76.64 | 69.79 | 53.60 |
| Boost-Soft | 17,123 | 9318 | 3533 | 64.76 | 82.90 | 72.71 | 57.13 |
| D2l | 16,039 | 9634 | 4617 | 62.47 | 77.65 | 69.24 | 52.95 |
| SCE | 15,099 | 7907 | 5567 | 65.63 | 73.06 | 69.15 | 52.84 |
| Peer | 15,896 | 10,532 | 4759 | 60.15 | 76.96 | 67.52 | 50.97 |
| DT-Forward | 13,787 | 5248 | 6869 | 72.43 | 66.75 | 69.47 | 53.22 |
| NCE-SCE | 14,319 | 7150 | 6337 | 66.70 | 69.32 | 67.98 | 51.50 |
| BaseLine_OSAMTL | 16,163 | 2230 | 4492 | 87.88 | 78.25 | 82.79 | 70.63 |

**Table 5.** (*Continued*).

| Solution | LAM | | | | | LMP | | |
|---|---|---|---|---|---|---|---|---|
| | LTP | LFP | LFN | Lprecision | Lrecall | Lf1 | | LfIoU |
| Forward_OSAMTL | | 16,197 | 2860 | 4459 | 84.99 | 78.41 | 81.57 | 68.88 |
| Backward_OSAMTL | | 16,167 | 3331 | 4489 | 82.92 | 78.27 | 80.52 | 67.4 |
| Boost-Hard_OSAMTL | | 16,560 | 2589 | 4095 | 86.48 | 80.17 | 83.21 | 71.24 |
| Boost-Soft_OSAMTL | | 15,778 | 2917 | 4878 | 84.4 | 76.38 | 80.19 | 66.93 |
| D2l_OSAMTL | | 16,108 | 2074 | 4547 | 88.59 | 77.99 | 82.95 | 70.87 |
| SCE_OSAMTL | | 14,907 | 2961 | 5748 | 83.43 | 72.17 | 77.39 | 63.12 |
| Peer_OSAMTL | | 16,983 | 4091 | 3673 | 80.59 | 82.22 | 81.39 | 68.63 |
| DT-Forward_OSAMTL | | 15,927 | 2045 | 4729 | 88.62 | 77.11 | 82.46 | 70.16 |
| NCE-SCE_OSAMTL | | 15,540 | 1971 | 5116 | 88.74 | 75.23 | 81.43 | 68.68 |

## 5.3. Results of US-based evaluations with AGTLs

The results of US-based evaluations with AGTLs for various methods of learning from inaccurate labels for the tumour segmentation in HE-stained pre-treatment biopsy images and the tumour segmentation in HE-stained post-treatment surgical resection images are respectively shown in **Tables 6** and **7**.

**Table 6.** US-based evaluations with AGTLs on the task of tumour segmentation in HE-stained pre-treatment biopsy images.

| Solution | TP | FP | FN | precision | recall | f1 | fIoU |
|---|---|---|---|---|---|---|---|
| BaseLine | 22,707 | 13,298 | 3249 | 63.07 | 87.48 | 73.29 | 57.85 |
| Forward | 23,494 | 15,160 | 2462 | 60.78 | 90.51 | 72.73 | 57.14 |
| Backward | 21,858 | 13,453 | 4098 | 61.90 | 84.21 | 71.35 | 55.46 |
| Boost-Hard | 22,184 | 12,652 | 3771 | 63.68 | 85.47 | 72.98 | 57.46 |
| Boost-Soft | 23,724 | 15,849 | 2231 | 59.95 | 91.40 | 72.41 | 56.75 |
| D2l | 23,068 | 14,632 | 2888 | 61.19 | 88.87 | 72.48 | 56.83 |
| SCE | 22,753 | 13,499 | 3203 | 62.76 | 87.66 | 73.15 | 57.67 |
| Peer | 22,658 | 12,704 | 3298 | 64.07 | 87.29 | 73.90 | 58.61 |
| DT-Forward | 23,280 | 14,239 | 2676 | 62.05 | 89.69 | 73.35 | 57.92 |
| NCE-SCE | 23,395 | 14,452 | 2561 | 61.81 | 90.13 | 73.34 | 57.90 |
| BaseLine_OSAMTL | 21,010 | 6381 | 4946 | 76.70 | 80.94 | 78.77 | 64.97 |
| Forward_OSAMTL | 20,215 | 5579 | 5740 | 78.37 | 77.88 | 78.13 | 64.11 |
| Backward_OSAMTL | 20,818 | 6124 | 5137 | 77.27 | 80.21 | 78.71 | 64.9 |
| Boost-Hard_OSAMTL | 20,230 | 5732 | 5725 | 77.92 | 77.94 | 77.93 | 63.84 |
| Boost-Soft_OSAMTL | 20,657 | 5936 | 5298 | 77.68 | 79.59 | 78.62 | 64.77 |
| D2l_OSAMTL | 20,348 | 5981 | 5608 | 77.28 | 78.39 | 77.83 | 63.71 |
| SCE_OSAMTL | 19,719 | 5651 | 6236 | 77.73 | 75.97 | 76.84 | 62.39 |
| Peer_OSAMTL | 20,379 | 6634 | 5577 | 75.44 | 78.51 | 76.95 | 62.53 |
| DT-Forward_OSAMTL | 19,958 | 5347 | 5998 | 78.87 | 76.89 | 77.87 | 63.76 |
| NCE-SCE_OSAMTL | 18,712 | 4594 | 7244 | 80.29 | 72.09 | 75.97 | 61.25 |

**Table 7.** US-based evaluations with AGTLs on the task of tumour segmentation in HE-stained post-treatment surgical resection images.

| Solution | TP | FP | FN | precision | recall | f1 | fIoU |
|---|---|---|---|---|---|---|---|
| BaseLine | 15,446 | 13,831 | 8467 | 52.76 | 64.59 | 58.08 | 40.92 |
| Forward | 15,129 | 13,409 | 8783 | 53.01 | 63.27 | 57.69 | 40.54 |
| Backward | 16,373 | 17,083 | 7540 | 48.94 | 68.47 | 57.08 | 39.94 |
| Boost-Hard | 16,599 | 15,904 | 7313 | 51.07 | 69.42 | 58.85 | 41.69 |
| Boost-Soft | 19,000 | 18,353 | 4912 | 50.87 | 79.46 | 62.03 | 44.95 |
| D2l | 16,331 | 14,876 | 7581 | 52.33 | 68.30 | 59.26 | 42.10 |
| SCE | 15,604 | 13,286 | 8309 | 54.01 | 65.25 | 59.10 | 41.95 |
| Peer | 17,366 | 19,348 | 6546 | 47.30 | 72.62 | 57.29 | 40.14 |
| DT-Forward | 15,374 | 15,525 | 8538 | 49.76 | 64.29 | 56.10 | 38.98 |
| NCE-SCE | 16,356 | 16,574 | 7556 | 49.67 | 68.40 | 57.55 | 40.40 |
| BaseLine_OSAMTL | 16,000 | 5649 | 7912 | 73.91 | 66.91 | 70.24 | 54.13 |
| Forward_OSAMTL | 14,825 | 3948 | 9088 | 78.97 | 62.00 | 69.46 | 53.21 |
| Backward_OSAMTL | 15,441 | 5648 | 8471 | 73.22 | 65.57 | 68.62 | 52.24 |
| Boost-Hard_OSAMTL | 15,713 | 4611 | 8200 | 77.31 | 65.71 | 71.04 | 55.09 |
| Boost-Soft_OSAMTL | 15,799 | 6017 | 8114 | 72.42 | 66.07 | 69.10 | 52.79 |
| D2l_OSAMTL | 15,109 | 3599 | 8803 | 80.76 | 63.18 | 70.90 | 54.92 |
| SCE_OSAMTL | 15,168 | 5151 | 8744 | 74.65 | 63.43 | 68.59 | 52.19 |
| Peer_OSAMTL | 16,954 | 7478 | 6958 | 69.39 | 70.90 | 70.14 | 54.01 |
| DT-Forward_OSAMTL | 15,175 | 4483 | 8737 | 77.20 | 63.46 | 69.66 | 53.44 |
| NCE-SCE_OSAMTL | 13,101 | 2749 | 10,811 | 82.66 | 54.79 | 65.90 | 49.14 |

## 5.4. Comparison between LAF and US

**Table 8.** Results for LAF-based evaluations (Lf1 and LfIoU) and US-based evaluations (f1 and fIoU) on easier task.

| Solution (Metric) Mean (CI) | SotA (Lf1) 78.91 (76.36–81.46) | SotA (LfIoU) 65.23 (61.83–68.63) | SotA (f1) 72.90 (72.23–73.57) | SotA (fIoU) 57.36 (56.53–58.19) |
|---|---|---|---|---|
| SotA-OSAMTL(Lf1) 78.04 (76.78–79.29) | $P = 0.372$ | | | |
| SotA-OSAMTL(LfIoU) 64.00 (62.32–65.68) | | $P = 0.343$ | | |
| SotA-OSAMTL(f1) 77.76 (76.89–78.63) | | | $P < 0.001$ | |
| SotA-OSAMTL (fIoU) 63.62 (62.46–64.78) | | | | $P < 0.001$ |

For the comparison between LAF and US, we compute the mean values with corresponding confident intervals (CI) and the P values of the overall performances for the state-of-the-art methods (SotA) and SotA combined with the improved OSAMTL (SotA-OSAMTL). The results for LAF-based evaluations with IAGTLs (Lf1 and LfIoU) and US-based evaluations with AGTLs (f1 and fIoU) on the task of tumour segmentation in HE-stained pre-treatment biopsy images (i.e., easier task) are shown in **Table 8**. The results for LAF-based evaluations with IAGTLs (Lf1 and LfIoU) and US-based evaluations with AGTLs (f1 and fIoU) on the task of tumour segmentation in HE-stained post-treatment surgical resection images (i.e., a more difficult task) are shown in **Table 9**.

**Table 9.** Results for LAF-based evaluations (Lf1 and LfIoU) and US-based evaluations (f1 and fIoU) on more difficult task.

| Solution (Metric) Mean (CI) | SotA (Lf1) 69.53(67.88–71.19) | SotA (LfIoU) 53.32(51.36–55.28) | SotA (f1) 58.30(56.75–59.86) | SotA (fIoU) 41.16(39.60–42.72) |
|---|---|---|---|---|
| SotA-OSAMTL(Lf1) 81.39(79.74–83.04) | $P < 0.001$ | | | |
| SotA-OSAMTL(LfIoU) 68.65(66.35–70.96) | | $P < 0.001$ | | |
| SotA-OSAMTL(f1) 69.37(67.96–70.77) | | | $P < 0.001$ | |
| SotA-OSAMTL (fIoU) 53.12(51.48–54.75) | | | | $P < 0.001$ |

## 5.5. Analysis

From **Table 8**, we can summarise that, on the easier task, the results of US-based evaluations with AGTLs (f1 and fIoU) show the advantages of the SotA-OSAMTL series compared with the SotA series (f1: $P < 0.001$, fIoU: $P < 0.001$), while the results of LAF-based evaluations with IAGTLs (Lf1 and LfIoU) do not show the same conclusions (Lf1: $P = 0.372$, LfIoU: $P = 0.343$). Since the previous work [11] has confirmed the advantages of the improved OSAMTL series compared with the state-of-the-art series [12–19] using US-based evaluations with AGTLs, the summarization from **Table 8** indicates that evaluations of LAF with IAGTLs cannot show the advantages of the SotA-OSAMTL series compared with the StoA series, just being unable to confidently act like evaluations of US with AGTLs on the easier task.

From **Table 9**, we can summarise that, on the more difficult task, the results of US-based evaluations with AGTLs (f1 and fIoU) show the advantages of the SotA-OSAMTL series compared with the SotA series (f1: $P < 0.001$, fIoU: $P < 0.001$), while the results of LAF-based evaluations with IAGTLs (Lf1 and LfIoU) as well show the same conclusions (Lf1: $P < 0.001$, LfIoU: $P < 0.001$). Identically, since the previous work [11] has confirmed the advantages of the improved OSAMTL series compared with the state-of-the-art series [12–19] using US-based evaluations with AGTLs, the summarization from **Table 9** indicates that evaluations of LAF with IAGTLs can show the advantages of the SotA-OSAMTL series compared with the StoA series, just being able to reasonably act like evaluations of US with AGTLs on the more difficult task.

As a result, the summarizations from **Tables 8** and **9** reflect that the practicability of LAF for evaluations with IAGTLs is valid in the case of TSfBC in MHWSIA.

## 6. Conclusion and discussion

In this paper, we validate the practicability of the logical assessment formula (LAF) for evaluations with inaccurate ground-truth labels (IAGTLs). The practicability of LAF for evaluations with IAGTLs includes: 1) LAF can be applied for evaluations with IAGTLs on a more difficult task, able to act like usual strategies for evaluations with AGTLs reasonably; and 2) LAF can be applied for evaluations with IAGTLs simply from the logical point of view on an easier task, unable to act like usual strategies for evaluations with AGTLs confidently. We applied LAF to two tasks of tumour segmentation for breast cancer (TSfBC) in medical histopathology whole slide image analysis (MHWSIA), and implemented a specific LAF solution that is suitable for evaluations with IAGTLs in the case of TSfBC in MHWSIA. Experimental results and analyses of this application support that the practicability of LAF for evaluations with IAGTLs is valid in the case of TSfBC in MHWSIA. Thus, the primary significance of this paper is that it reports a positive study that reflects the potential of LAF applied to MHWSIA for evaluations with IAGTLs. This paper presents the first practical validation of LAF for evaluations with IAGTLs in a real-world application.

Although the application of LAF to TSfBC in MHWSIA showed good support for the practicability of LAF, the problem that remains unsolved is how to estimate whether a given task is a difficult one or an easy one in the application of LAF for evaluations without AGTL. Since the practicability of LAF reflects that evaluations of LAF with IAGTLs on a difficult task are more reliable (more consistent with evaluations of usual strategies with AGTL) than on an easier task, the definition of a given task as difficult or easy is the key foundation for the application of LAF for evaluations with IAGTL. In this paper, the estimation of the two tasks of TSfBC in MHWSIA to be difficult or easy is qualitatively formed by the problem analyses and suggestions from pathology experts [11] (Section 4.1), and fortunately, the two tasks are suitable to validate the practicability of LAF. This specific validation demonstrates the practicability of LAF is valid with the case of TSfBC in MHWSIA, but it is not persuasive enough to help deciding whether LAF is suitable for evaluations IAGTL on any other given task. However, if the difficulty of a given task can be quantitatively estimated, then it will be much easier for us to decide whether LAF is suitable for evaluations with IAGTL on the given task via an appropriate threshold of task difficulty. Moreover, more applications of LAF applied to other tasks need to be conducted. In future works, these issues should be addressed.

**Supplementary materials:** Detailed proofs for the reasoning results presented in this article are provided in the supplementary materials.

**Author contributions:** Conceptualization, YY; methodology, YY; software, YY; validation, YY and HB; formal analysis, YY and HB; investigation, YY; resources, HB and YY; data curation, HB and YY; writing—original draft preparation, YY; writing—review and editing, HB; visualization, YY; supervision, YY and HB; project administration, YY and HB; funding acquisition, HB. All authors have read and agreed to the published version of the manuscript.

**Acknowledgments:** We acknowledge Yani Wei and Fengling Li for providing the annotations for the data used for experiments when they were PhD candidates supervised by Hong Bu, and Zhongjiu Flash Medical Technology Co., Ltd., Mianyang, China for providing the technical supports for revisions of this paper.

**Funding:** This work was supported by the 1·3·5 project for disciplines of excellence (ZYGD18012); the Technological Innovation Project of Chengdu New Industrial

Technology Research Institute (2017-CY02–00026-GX).

**Competing interest:** The authors declare no conflict of interest.

# Reference

1. Yang Y. Logical assessment formula and its principles for evaluations with inaccurate ground-truth labels. Knowledge and Information Systems. 2024; 66(4): 2561–2573. doi: 10.1007/s10115-023-02047-6

2. Chang HH, Zhuang AH, Valentino DJ, et al. Performance measure characterization for evaluating neuroimage segmentation algorithms. NeuroImage. 2009; 47(1): 122–135. doi: 10.1016/j.neuroimage.2009.03.068

3. Taha AA, Hanbury A. Metrics for evaluating 3D medical image segmentation: analysis, selection, and tool. BMC Medical Imaging. 2015; 15(1). doi: 10.1186/s12880-015-0068-x

4. M H, M.N S. A Review on Evaluation Metrics for Data Classification Evaluations. International Journal of Data Mining & Knowledge Management Process. 2015; 5(2): 01–11. doi: 10.5121/ijdkp.2015.5201

5. Jung HJ, Lease M. Evaluating Classifiers Without Expert Labels. Published online 2012. doi: 10.48550/ARXIV.1212.0960

6. Deng W, Zheng L. Are Labels Always Necessary for Classifier Accuracy Evaluation? 2021 IEEE/CVF Conference on Computer Vision and Pattern Recognition (CVPR). Published online June 2021. doi: 10.1109/cvpr46437.2021.01482

7. Joyce RJ, Raff E, Nicholas C. A Framework for Cluster and Classifier Evaluation in the Absence of Reference Labels. Proceedings of the 14th ACM Workshop on Artificial Intelligence and Security. Published online November 15, 2021. doi: 10.1145/3474369.3486867

8. Bouix S, Martin-Fernandez M, Ungar L, et al. On evaluating brain tissue classifiers without a ground truth. NeuroImage. 2007; 36(4): 1207–1224. doi: 10.1016/j.neuroimage.2007.04.031

9. Warfield SK, Zou KH, Wells WM. Simultaneous Truth and Performance Level Estimation (STAPLE): An Algorithm for the Validation of Image Segmentation. IEEE Transactions on Medical Imaging. 2004; 23(7): 903–921. doi: 10.1109/tmi.2004.828354

10. Martin-Fernandez M, Bouix S, Ungar L, et al. Two Methods for Validating Brain Tissue Classifiers. In: Duncan JS, Gerig G. (editors). Medical Image Computing and Computer-Assisted Intervention—MICCAI 2005. Springer Berlin Heidelberg; 2005. pp 515–522.

11. Yang Y, Li F, Wei Y, et al. One-step abductive multi-target learning with diverse noisy samples and its application to tumour segmentation for breast cancer. Expert Systems with Applications. 2024; 251: 123923. doi: 10.1016/j.eswa.2024.123923

12. Patrini G, Rozza A, Menon AK, et al. Making Deep Neural Networks Robust to Label Noise: A Loss Correction Approach. 2017 IEEE Conference on Computer Vision and Pattern Recognition (CVPR). Published online July 2017. doi: 10.1109/cvpr.2017.240

13. Reed SE, Lee H, Anguelov D, et al. Training deep neural networks on noisy labels with bootstrapping. In: Proceeding of 3rd International Conference on Learning Representations, ICLR 2015 - Workshop Track Proceedings. 2015.

14. Arazo E, Ortego D, Albert P, et al. Unsupervised label noise modeling and loss correction. In: 36th International Conference on Machine Learning; 2019.

15. Ma X, Wang Y, Houle ME, et al. Dimensionality-Driven learning with noisy labels. In: 35th International Conference on Machine Learning, ICML; 2018.

16. Wang Y, Ma X, Chen Z, et al. Symmetric Cross Entropy for Robust Learning With Noisy Labels. 2019 IEEE/CVF International Conference on Computer Vision (ICCV). Published online October 2019. doi: 10.1109/iccv.2019.00041

17. Liu Y, Guo H. Peer loss functions: Learning from noisy labels without knowing noise rates. In: 37th International Conference on Machine Learning; 2020.

18. Yao Y, Liu T, Han B, et al. Dual T: Reducing estimation error for transition matrix in label-noise learning. In: Advances in Neural Information Processing Systems; 2020.

19. Ma X, Huang H, Wang Y, et al. Normalized loss functions for deep learning with noisy labels. In: Processing of 37th International Conference on Machine Learning; 2020.

20. Yang Y, Yang Y, Yuan Y, et al. Detecting helicobacter pylori in whole slide images via weakly supervised multi-task learning. Multimedia Tools and Applications. 2020; 79(35–36): 26787–26815. doi: 10.1007/s11042-020-09185-x

21. Yang Y, Yang Y, Chen J, et al. Handling noisy labels via one-step abductive multi-target learning and its application to helicobacter pylori segmentation. Multimedia Tools and Applications. 2024; 83(24): 65099–65147. doi: 10.1007/s11042-

023-17743-2

22. Yang Y. Discovering Scientific Paradigms for Artificial Intelligence Alignment. 2023. doi: 10.13140/RG.2.2.15945.52320

23. Frenay B, Verleysen M. Classification in the Presence of Label Noise: A Survey. IEEE Transactions on Neural Networks and Learning Systems. 2014; 25(5): 845–869. doi: 10.1109/tnnls.2013.2292894

24. Song H, Kim M, Park D, et al (2020) Learning from Noisy Labels with Deep Neural Networks: A Survey

25. Long J, Shelhamer E, Darrell T. Fully convolutional networks for semantic segmentation. 2015 IEEE Conference on Computer Vision and Pattern Recognition (CVPR). Published online June 2015. doi: 10.1109/cvpr.2015.7298965

Article

# Validation of the practicability of logical assessment formula for evaluations with inaccurate ground-truth labels: An application study on tumour segmentation for breast cancer

## Supplementary materials

## Preliminary of logical reasoning

We introduce some propositional connectives and rules for proof of propositional logical reasoning, which are respectively shown as **Table S1** and **Table S2**, for the logical reasonings conducted in this paper.

**Table S1.** Propositional connectives.

| Connective | Meaning |
|---|---|
| ∧ | Conjunction |
| → | Implication |

**Table S2.** Rules for proof of propositional logical reasoning, ⊢ denotes 'bring out'.

| Rule | Meaning |
|---|---|
| ∧ − | Reductive law of conjunction: A ∧ B, ⊢ A or B. |
| ∧ + | Additional law of conjunction: A, B, ⊢ A ∧ B. |
| MP | Modus ponens: A → B, A, ⊢ B. |
| HS | Hypothetical syllogism: A → B, B → C, ⊢ A → C. |

## Proof of Reasoning 1

**Reasoning 1.** *If $\tilde{t}_{TSfBC,1}$ is given, then pixels included in negative areas of $\tilde{t}_{TSfBC,1}$ are most probably true tumour negatives.*

**Proof.** Firstly, with the given $\tilde{t}_{TSfBC,1}$, we have following preconditions for Reasoning 1.

1) If $\tilde{t}_{TSfBC,1}$ is given, then the recall of positive areas of $\tilde{t}_{TSfBC,1}$ to represent true tumour positives is very high.
2) If the recall of positive areas of $\tilde{t}_{TSfBC,1}$ to represent true tumour positives is very high, then almost all of true tumour positives are included in positive areas of $\tilde{t}_{TSfBC,1}$.
3) If almost all of true tumour positives are included in positive areas of $\tilde{t}_{TSfBC,1}$, then true tumour positives included in negative areas of $\tilde{t}_{TSfBC,1}$ are rare.
4) If true tumour positives included in negative areas of $\tilde{t}_{TSfBC,1}$ are rare, then pixels included in negative areas of $\tilde{t}_{TSfBC,1}$ are mostly probably true tumour negatives.

Secondly, we give the propositional symbols for the above preconditions 1–4 for Reasoning 1, which are shown in **Table S3**.

**Table S3.** Propositional symbols of preconditions for Reasoning 1.

| Symbol | Meaning |
|--------|---------|
| $a$ | $\tilde{t}_{TSfBC,1}$ is given. |
| $b$ | The recall of positive areas of $\tilde{t}_{TSfBC,1}$ to represent true tumour positives is very high. |
| $c$ | Almost all of true tumour positives are included in positive areas of $\tilde{t}_{TSfBC,1}$. |
| $d$ | True tumour positives included in negative areas of $\tilde{t}_{TSfBC,1}$ are rare |
| $e$ | Pixels included in negative areas of $\tilde{t}_{TSfBC,1}$ are mostly probably true tumour negatives |

Thirdly, referring to **Table S3**, we signify the propositional formalizations of the preconditions 1–4 for Reasoning 1 and Reasoning 1 via the propositional connectives listed in **Table S1** as follows.

1) $a \rightarrow b$                            Precondition
2) $b \rightarrow c$                            Precondition
3) $c \rightarrow d$                            Precondition
4) $d \rightarrow e$                            Precondition
    $a \rightarrow e$                            Reasoning 1

Fourthly, we show the validity of Reasoning 1 via the rules for proof of propositional logical reasoning listed in **Table S2** as follows.

$\therefore \boldsymbol{a \rightarrow e}$

5) $a$                                Hypothesis
6) $a \rightarrow c$                           1),2); HS
7) $c \rightarrow e$                           3),4); HS
8) $a \rightarrow e$                           6),7); HS
9) $e$                                8),5); MP
10) $a \rightarrow e$                       5)-9); Conditional Proof

Since the hypothesis $a$ of the 5) step has been fulfilled by the abduced $\tilde{t}_{TSfBC} = \{\tilde{t}_{TSfBC,1}, \tilde{t}_{TSfBC,2}\}$ in section 5.2.2., Reasoning 1 is proved to be valid. □

**Proof of Reasoning 2**

**Reasoning 2.** *If $\tilde{t}_{TSfBC,2}$ is given, then pixels included in positive areas of $\tilde{t}_{TSfBC,2}$ are most probably true tumour positives.*

**Proof.** Firstly, with the given $\tilde{t}_{TSfBC,2}$, we have following preconditions for Reasoning 2.

1) If $\tilde{t}_{TSfBC,2}$ is given, then the precision of positive areas of $\tilde{t}_{TSfBC,2}$ to represent true tumour positives is very high.
2) If the precision of positive areas of $\tilde{t}_{TSfBC,2}$ to represent true tumour positives is very high, then the positive areas of $\tilde{t}_{TSfBC,2}$ are almost all true tumour positives.
3) If the positive areas of $\tilde{t}_{TSfBC,2}$ are almost all true tumour positives, then false tumour positives included in positive areas of $\tilde{t}_{TSfBC,2}$ are rare.
4) If false tumour positives included in positive areas of $\tilde{t}_{TSfBC,2}$ are rare, then pixels included in positive areas of $\tilde{t}_{TSfBC,2}$ are most probably true tumour positives.

Secondly, we give the propositional symbols for the above preconditions 1–4 for Reasoning 2, which are shown in **Table S4**.

**Table S4.** Propositional symbols of preconditions for Reasoning 2.

| Symbol | Meaning |
|--------|---------|
| $f$ | $\tilde{t}_{TSfBC,2}$ is given |
| $g$ | The precision of positive areas of $\tilde{t}_{TSfBC,2}$ to represent true tumour positives is very high. |
| $h$ | The positive areas of $\tilde{t}_{TSfBC,2}$ are almost all true tumour positives. |
| $i$ | False tumour positives included in positive areas of $\tilde{t}_{TSfBC,2}$ are rare. |
| $j$ | Pixels included in positive areas of $\tilde{t}_{TSfBC,2}$ are most probably true tumour positives. |

Thirdly, referring to **Table S4**, we signify the propositional formalizations of the preconditions 1–4 for Reasoning 2 and Reasoning 2 via the propositional connectives listed in **Table S1** as follows.

1) $f \rightarrow g$                        Precondition
2) $g \rightarrow h$                        Precondition
3) $h \rightarrow i$                         Precondition
4) $i \rightarrow j$                         Precondition
    $f \rightarrow j$                         Reasoning 2

Fourthly, we show the validity of Reasoning 2 via the rules for proof of propositional logical reasoning listed in **Table S2** as follows.

$$\therefore f \rightarrow j$$

5) $f$                            Hypothesis
6) $f \rightarrow h$                       1), 2); HS
7) $h \rightarrow j$                       3), 4); HS
8) $f \rightarrow j$                       6), 7); HS
9) $j$                            8), 5); MP
10) $f \rightarrow j$                     5)–9); Conditional Proof

Since the hypothesis $f$ of the 5) step has been fulfilled by the abduced $\tilde{t}_{TSfBC} = \{\tilde{t}_{TSfBC,1}, \tilde{t}_{TSfBC,2}\}$ in section 5.2.2., Reasoning 2 is proved to be valid. □

**Proof of Reasoning 3**

**Reasoning 3.** *If $t_{TSfBC}$ is given and $LF_{TSfBC,1}$ is given, then the intersection of pixels of $t_{TSfBC}$ that are predicted as tumour positives ($t_{TSfBC}^p$) and pixels included in negative areas of $\tilde{t}_{TSfBC,1}$ ($\tilde{t}_{TSfBC,1}^n$) can be considered as logically false positives.*

**Proof.** Firstly, with the given $t_{TSfBC}$ and $LF_{TSfBC,1}$, we have following preconditions for Reasoning 3.

1) If $LF_{TSfBC,1}$ is given, then $\tilde{t}_{TSfBC,1}$ is given.

2) If $\tilde{t}_{TSfBC,1}$ is given, then pixels included in negative areas of $\tilde{t}_{TSfBC,1}$ ($\tilde{t}_{TSfBC,1}^n$) are most probably true tumour negatives. (Reasoning 1)

3) If $t_{TSfBC}$ is given, then pixels of $t_{TSfBC}$ that are predicted as tumour positives ($t_{TSfBC}^p$) exist.

4) If pixels included in negative areas of $\tilde{t}_{TSfBC,1}$ ($\tilde{t}_{TSfBC,1}^n$) are most probably true tumour negatives and pixels of $t_{TSfBC}$ that are predicted as tumour positives ($t_{TSfBC}^p$) exist, then the intersection of pixels included in $t_{TSfBC}^p$ and pixels included in $\tilde{t}_{TSfBC,1}^n$ can be considered as most probably predicted false tumour positives.

5) If the intersection of pixels included in $t_{TSfBC}^p$ and pixels included in $\tilde{t}_{TSfBC,1}^n$ can be considered as most probably predicted false tumour positives, then the intersection of pixels included in $t_{TSfBC}^p$ and pixels included in $\tilde{t}_{TSfBC,1}^n$ can be considered as logically false positives.

6) If the intersection of pixels included in $t_{TSfBC}^p$ and pixels included in $\tilde{t}_{TSfBC,1}^n$ can be considered as logically false positives, then the intersection of pixels of $t_{TSfBC}$ that are predicted as tumour positives ($t_{TSfBC}^p$) and pixels included in negative areas of $\tilde{t}_{TSfBC,1}$ ($\tilde{t}_{TSfBC,1}^n$) can be considered as logically false positives.

Secondly, we give the propositional symbols for the above preconditions 1–6 for Reasoning 3, which are shown in **Table S5**.

**Table S5.** Propositional symbols of preconditions for Reasoning 3.

| Symbol | Meaning |
|---|---|
| $k$ | $LF_{TSfBC,1}$ is given. |
| $l$ | $\tilde{t}_{TSfBC,1}$ is given. |
| $m$ | Pixels included in negative areas of $\tilde{t}_{TSfBC,1}$ ($\tilde{t}_{TSfBC,1}^n$) are most probably true tumour negatives. |
| $n$ | $t_{TSfBC}$ is given. |
| $o$ | Pixels of $t_{TSfBC}$ that are predicted as tumour positives ($t_{TSfBC}^p$) exist. |
| $p$ | The intersection of pixels included in $t_{TSfBC}^p$ and pixels included in $\tilde{t}_{TSfBC,1}^n$ can be considered as most probably predicted false tumour positives. |
| $q$ | The intersection of pixels included in $t_{TSfBC}^p$ and pixels included in $\tilde{t}_{TSfBC,1}^n$ can be considered as logically false positives. |
| $r$ | The intersection of pixels of $t_{TSfBC}$ that are predicted as tumour positives ($t_{TSfBC}^p$) and pixels included in negative areas of $\tilde{t}_{TSfBC,1}$ ($\tilde{t}_{TSfBC,1}^n$) can be considered as logically false positives. |

Thirdly, referring to **Table S5**, we signify the propositional formalizations of the preconditions 1–6 for Reasoning 3 and Reasoning 3 via the propositional connectives listed in **Table S1** as follows.

1) $k \rightarrow l$                           Precondition
2) $l \rightarrow m$                          Precondition
3) $n \rightarrow o$                          Precondition
4) $(m \wedge o) \rightarrow p$               Precondition
5) $p \rightarrow q$                          Precondition
6) $q \rightarrow r$                          Precondition
    $(n \wedge k) \rightarrow r$             Reasoning 3

Fourthly, we show the validity of Reasoning 3 via the rules for proof of propositional logical reasoning listed in **Table S2** as follows.

∴ $(\boldsymbol{n \wedge k}) \rightarrow \boldsymbol{r}$

7) $n \wedge k$                          Hypothesis
8) $n$                              7); $\wedge -$
9) $k$                              7); $\wedge -$
10) $l$                             1), 9); MP
11) $m$                           2), 10); MP
12) $o$                             3), 8); MP
13) $m \wedge o$                       11), 12); $\wedge +$
14) $(m \wedge o) \rightarrow q$          4), 5); HS
15) $(m \wedge o) \rightarrow r$          14), 6); HS
16) $r$                            15), 13); MP
17) $(n \wedge k) \rightarrow r$            7)–16); Conditional Proof

Since the hypothesis $n \wedge k$ of the 7) step has been fulfilled by the prediction of the image semantic segmentation model for tumour segmentation for breast cancer ($t_{TSfBC}$) in section 5.2.3. and the two narrated logical facts $LF_{TSfBC} = \{LF_{TSfBC,1}, LF_{TSfBC,2}\}$, Reasoning 3 is proved to be valid. □

**Proof of Reasoning 4**

**Reasoning 4.** *If $t_{TSfBC}$ is given and $LF_{TSfBC,2}$ is given, then the intersection of pixels of $t_{TSfBC}$ that are predicted as tumour positives ($t^p_{TSfBC}$) and pixels included in positive areas of $\tilde{t}_{TSfBC,2}$ ($\tilde{t}^p_{TSfBC,2}$) can be considered as logically true positives, and the intersection of pixels of $t_{TSfBC}$ that are predicted as tumour negatives ($t^n_{TSfBC}$) and pixels included in positive areas of $\tilde{t}_{TSfBC,2}$ ($\tilde{t}^p_{TSfBC,2}$) can be considered as logically false negatives.*

**Proof.** Firstly, with the given $t_{TSfBC}$ and $LF_{TSfBC,2}$, we have following preconditions for Reasoning 4.

1)  If $LF_{TSfBC,2}$ is given, then $\tilde{t}_{TSfBC,2}$ is given.

2)  If $\tilde{t}_{TSfBC,2}$ is given, then pixels included in positive areas of $\tilde{t}_{TSfBC,2}$ ($\tilde{t}^p_{TSfBC,2}$) are most probably true tumour positives. (Reasoning 2).

3)  If $t_{TSfBC}$ is given, then pixels of $t_{TSfBC}$ that are predicted as tumour positives ($t^p_{TSfBC}$) exist and pixels of $t_{TSfBC}$ that are predicted as tumour negatives ($t^n_{TSfBC}$) exist.

4)  If pixels included in positive areas of $\tilde{t}_{TSfBC,2}$ ($\tilde{t}^p_{TSfBC,2}$) are most probably true tumour positives and pixels of $t_{TSfBC}$ that are predicted as tumour positives ($t^p_{TSfBC}$) exist, then the intersection of pixels included in $t^p_{TSfBC}$ and pixels included in $\tilde{t}^p_{TSfBC,2}$ can be considered as most probably predicted true tumour positives.

5)  If pixels included in positive areas of $\tilde{t}_{TSfBC,2}$ ($\tilde{t}^p_{TSfBC,2}$) are most probably true tumour positives and pixels of $t_{TSfBC}$ that are predicted as tumour negatives ($t^n_{TSfBC}$) exist, then the intersection of pixels included in $t^n_{TSfBC}$ and pixels included in $\tilde{t}^p_{TSfBC,2}$ can be considered as most probably predicted false tumour negatives.

6)  If the intersection of pixels included in $t^p_{TSfBC}$ and pixels included in $\tilde{t}^p_{TSfBC,2}$ can be considered as most probably predicted true tumour positives, then the intersection of pixels included in $t^p_{TSfBC}$ and pixels included in $\tilde{t}^p_{TSfBC,2}$ can be considered as logically true positives.

7)  If the intersection of pixels included in $t^n_{TSfBC}$ and pixels included in $\tilde{t}^p_{TSfBC,2}$ can be considered as most probably predicted false tumour negatives, then the intersection of pixels included in $t^n_{TSfBC}$ and pixels included in $\tilde{t}^p_{TSfBC,2}$ can be considered as logically false negatives.

8)  If the intersection of pixels included in $t^p_{TSfBC}$ and pixels included in $\tilde{t}^p_{TSfBC,2}$ can be considered as logically true positives, then the intersection of pixels of $t_{TSfBC}$ that are predicted as tumour positives ($t^p_{TSfBC}$) and pixels included in positive areas of $\tilde{t}_{TSfBC,2}$ ($\tilde{t}^p_{TSfBC,2}$) can be considered as logically true positives.

9)  If the intersection of pixels included in $t^n_{TSfBC}$ and pixels included in $\tilde{t}^p_{TSfBC,2}$ can be considered as logically false negatives, then the intersection of pixels of $t_{TSfBC}$ that are predicted as tumour negatives ($t^n_{TSfBC}$) and pixels included in positive areas of $\tilde{t}_{TSfBC,2}$ ($\tilde{t}^p_{TSfBC,2}$) can be considered as logically false negatives.

Secondly, we give the propositional symbols for the above preconditions 1-9 for Reasoning 4, which are shown in **Table S6**.

**Table S6.** Propositional symbols of preconditions for Reasoning 4.

| Symbol | Meaning |
|--------|---------|
| $s$ | $LF_{TSfBC,2}$ is given. |
| $t$ | $\tilde{t}_{TSfBC,2}$ is given. |
| $u$ | Pixels included in positive areas of $\tilde{t}_{TSfBC,2}$ ($\tilde{t}^p_{TSfBC,2}$) are most probably true tumour positives. |
| $v$ | $t_{TSfBC}$ is given. |
| $w$ | Pixels of $t_{TSfBC}$ that are predicted as tumour positives ($t^p_{TSfBC}$) exist. |
| $x$ | Pixels of $t_{TSfBC}$ that are predicted as tumour negatives ($t^n_{TSfBC}$) exist. |
| $y$ | The intersection of pixels included in $t^p_{TSfBC}$ and pixels included in $\tilde{t}^p_{TSfBC,2}$ can be considered as most probably predicted true tumour positives. |
| $z$ | The intersection of pixels included in $t^n_{TSfBC}$ and pixels included in $\tilde{t}^p_{TSfBC,2}$ can be considered as most probably predicted false tumour negatives. |
| $a$ | The intersection of pixels included in $t^p_{TSfBC}$ and pixels included in $\tilde{t}^p_{TSfBC,2}$ can be considered as logically true positives. |
| $b$ | The intersection of pixels included in $t^n_{TSfBC}$ and pixels included in $\tilde{t}^p_{TSfBC,2}$ can be considered as logically false negatives. |
| $c$ | The intersection of pixels of $t_{TSfBC}$ that are predicted as tumour positives ($t^p_{TSfBC}$) and pixels included in positive areas of $\tilde{t}_{TSfBC,2}$ ($\tilde{t}^p_{TSfBC,2}$) can be considered as logically true positives. |
| $d$ | The intersection of pixels of $t_{TSfBC}$ that are predicted as tumour negatives ($t^n_{TSfBC}$) and pixels included in positive areas of $\tilde{t}_{TSfBC,2}$ ($\tilde{t}^p_{TSfBC,2}$) can be considered as logically false negatives. |

Thirdly, referring to **Table S6**, we signify the propositional formalizations of the preconditions 1–9 for Reasoning 4 and Reasoning 4 via the propositional connectives listed in **Table S1** as follows.

1) $s \rightarrow t$          Precondition
2) $t \rightarrow u$          Precondition
3) $v \rightarrow (w \wedge x)$          Precondition
4) $(u \wedge w) \rightarrow y$          Precondition
5) $(u \wedge x) \rightarrow z$          Precondition
6) $y \rightarrow a$          Precondition
7) $z \rightarrow b$          Precondition
8) $a \rightarrow c$          Precondition
9) $b \rightarrow d$          Precondition
    $(v \wedge s) \rightarrow (c \wedge d)$          Reasoning 4

Fourthly, we show the validity of Reasoning 4 via the rules for proof of propositional logical reasoning listed in **Table S2** as follows.

∴ $\boldsymbol{(v \wedge s) \rightarrow (c \wedge d)}$

10) $v \wedge s$          Hypothesis
11) $v$          10); $\wedge -$
12) $s$          10); $\wedge -$
13) $s \rightarrow u$          1), 2); HS
14) $u$          13), 12); MP
15) $w \wedge x$          3), 11); MP
16) $w$          15); $\wedge -$

| | |
|---|---|
| 17) $x$ | 15); $\wedge -$ |
| 18) $u \wedge w$ | 14), 16); $\wedge +$ |
| 19) $(u \wedge w) \rightarrow a$ | 4), 6); HS |
| 20) $u \wedge x$ | 14), 17); $\wedge +$ |
| 21) $(u \wedge x) \rightarrow b$ | 5), 7); HS |
| 22) $(u \wedge w) \rightarrow c$ | 19), 8); HS |
| 23) $(u \wedge x) \rightarrow d$ | 21), 9); HS |
| 24) $c$ | 22), 18); MP |
| 25) $d$ | 23), 20); MP |
| 26) $c \wedge d$ | 24), 25); $\wedge +$ |
| 27) $(v \wedge s) \rightarrow (c \wedge d)$ | 10)–26); Conditional Proof |

Since the hypothesis $v \wedge s$ of the 10) step has been fulfilled by the prediction of the image semantic segmentation model for tumour segmentation for breast cancer ($t_{TSfBC}$) in section 5.2.3. and the two narrated logical facts $LF_{TSfBC} = \{LF_{TSfBC,1}, LF_{TSfBC,2}\}$ Reasoning 4 is proved to be valid. $\square$